\documentclass[11pt,a4paper]{article}
\usepackage[hyperref]{acl2018}
\usepackage{times}
\usepackage{latexsym}

\usepackage{url}	

\usepackage{amsmath}
\usepackage{amssymb}
\usepackage{multirow}
\usepackage{bm}
\usepackage{graphicx}
\usepackage{algorithmic}
\usepackage{algorithm}
\usepackage{etoolbox}
\usepackage{dsfont}
\usepackage{caption}
\usepackage{footmisc}

\usepackage{subfig}

\newcommand{\DP}[2]{{#1}^{\top}{#2}}

\aclfinalcopy 
\def\aclpaperid{1146} 

\newcommand{\bs}[1]{\bm{#1}}

\DeclareMathAlphabet{\mathbb}{U}{msb}{m}{n}
\DeclareMathOperator*{\softmax}{\mathsf{softmax}}
\DeclareMathOperator*{\sparsemax}{\mathsf{sparsemax}}
\DeclareMathOperator*{\csoftmax}{\mathsf{csoftmax}}
\DeclareMathOperator*{\csparsemax}{\mathsf{csparsemax}}
\DeclareMathOperator*{\argmin}{\mathsf{arg\,min}}

\DeclareMathOperator*{\embed}{\mathsf{embed}}

\DeclareMathOperator*{\rnn}{\mathsf{RNN}}

\newtheorem{theorem}{Theorem}

\newtheorem{proposition}[theorem]{\bf{Proposition}}

\newenvironment{itemizesquish}{\begin{list}{\labelitemi}{\setlength{\topsep}{0em}\setlength{\itemsep}{0em}\setlength{\labelwidth}{0.75em}\setlength{\leftmargin}{\labelwidth}\addtolength{\leftmargin}{\labelsep}}}{\end{list}}

\newcommand{\correct}[1]{{\color{blue}{\it #1}}}
\newcommand{\wrong}[1]{{\color{red}{\bf #1}}}
\newcommand{\source}[1]{{\textcolor{olive}{#1}}}

\title{Sparse and Constrained Attention for Neural Machine Translation}

\author{Chaitanya Malaviya\thanks{\,\,\,Work done during an internship at Unbabel.} \\
  Language Technologies Institute \\ Carnegie Mellon University \\
  {\tt cmalaviy@cs.cmu.edu} \\\And
  Pedro Ferreira \\
  Instituto Superior T\'ecnico \\ Universidade de Lisboa \\
  Lisbon, Portugal \\
  {\tt pedro.miguel.ferreira}\\
  {\tt @tecnico.ulisboa.pt} \\\And
  Andr\'e~F.~T. Martins   \\
  Unbabel \\ \& Instituto de Telecomunica\c{c}\~{o}es \\
  Lisbon, Portugal \\
  {\tt andre.martins} \\
  {\tt @unbabel.com}}

\date{}

\begin{document}
\maketitle
\begin{abstract}
In NMT, words are sometimes dropped from the source or  generated repeatedly in the translation.  
We explore novel strategies to address the coverage problem that 
change only the attention transformation.
Our approach allocates fertilities to source words, used to bound the attention each word can receive. 
We experiment with various sparse and constrained attention transformations and propose a new one, constrained sparsemax, 
shown to be differentiable and sparse. 
Empirical evaluation is
provided in three languages pairs.
\end{abstract}

\section{Introduction}


Neural machine translation (NMT) emerged in the last few years as a very successful paradigm   \citep{
Sutskever2014,bahdanau2014neural,Gehring2017,Vaswani2017}. 
While NMT is generally more fluent than previous statistical 
systems,
{adequacy} is still a major concern \cite{koehn-knowles:2017:NMT}:  
common mistakes include dropping source words and 
repeating words in the generated translation. 

Previous work has attempted to mitigate this problem 
in various ways. \citet{wu2016google} incorporate coverage and length penalties during beam search---%
a simple yet limited solution, since it only affects 
the scores of translation hypotheses that are already in the beam. 
Other approaches involve architectural changes: 
providing coverage vectors to track the attention history \cite{mi2016coverage, Tu:2016:ACL}, 
using gating architectures and adaptive attention to control the amount of source context provided \cite{Tu:2017:TACL,li2017learning}, or adding a reconstruction loss \cite{tu2017neural}. \citet{feng2016} also use the notion of fertility implicitly in their proposed model. Their “fertility conditioned decoder” uses a coverage vector and an “extract gate” which are incorporated in the decoding recurrent unit, increasing the number of parameters.



In this paper, we propose a different solution that does not change the overall architecture, but only the {\bf attention transformation}. 
Namely, we 
replace the traditional softmax 
by other recently proposed transformations that either promote attention sparsity \citep{martins2016softmax} 
or upper bound the amount of attention a word can receive \citep{martins2017learning}. 
The bounds are determined by the fertility values of the source words. 
While these transformations have given encouraging results in various NLP problems, they have never been applied to NMT, to the best of our knowledge. 
Furthermore, we combine these two ideas and propose a novel attention transformation, {\bf constrained sparsemax}, which produces \emph{both} sparse and bounded attention weights, 
yielding a compact and interpretable set of alignments. 
While being in-between soft and hard alignments (Figure~\ref{fig:att_maps}), the constrained sparsemax transformation is end-to-end differentiable, hence amenable for training with gradient backpropagation. 
%

To sum up, our contributions are as follows:%
\footnote{Our software code is available at the OpenNMT fork \texttt{www.github.com/Unbabel/OpenNMT-py/tree/dev} and the running scripts at \texttt{www.github.com/Unbabel/\\
sparse\_constrained\_attention}.}

\smallskip

\begin{itemizesquish}
\item We formulate constrained sparsemax 
and 
derive efficient linear and sublinear-time algorithms for running 
forward and backward propagation. 
This transformation has two levels of sparsity: over time steps, and over the attended words at each step. 
\item We provide a detailed empirical comparison of various attention transformations, 
including softmax \citep{bahdanau2014neural}, sparsemax \citep{martins2016softmax}, 
constrained softmax \citep{martins2017learning}, 
and our newly proposed constrained sparsemax. 
We provide error analysis including two new metrics targeted at detecting coverage problems. 
\end{itemizesquish}





\section{Preliminaries}

Our underlying model architecture is a  standard attentional encoder-decoder \citep{bahdanau2014neural}. 
Let $x:=x_{1:J}$ and $y:=y_{1:T}$ denote 
the source and target sentences, respectively. 
We use a Bi-LSTM encoder to represent the source words as a matrix $\bs{H} := [\bs{h}_1, \ldots, \bs{h}_J] \in \mathbb{R}^{2D \times J}$. 
The conditional probability of the target sentence is given as 
\begin{equation}
\textstyle p(y\,\,|\,\,x) := \prod_{t=1}^T p(y_{t}\,\,|\,\,y_{1:(t-1)}, x),
\end{equation}
where $p(y_{t}\,\,|\,\,y_{1:(t-1)}, x)$ is computed by a softmax output layer that receives 
a decoder state $\bs{s}_t$ as input. 
This state is updated by an auto-regressive LSTM, 
$\bs{s}_t = \rnn(\embed(y_{t-1}), \bs{s}_{t-1} , \bs{c}_t)$,
where $\bs{c}_t$ is an input context vector. 
This vector is 
computed as $\bs{c}_t := \bs{H}\bs{\alpha}_{t}$, 
where $\bs{\alpha}_{t}$ is a probability distribution that represents the attention over the source words, commonly obtained as 
\begin{equation}\label{eq:softmax_attention}
\bs{\alpha}_{t} = \softmax(\bs{z}_t),
\end{equation}
where $\bs{z}_t \in \mathbb{R}^J$ is a vector of scores. 
We follow \citet{Luong2015} and define 
$z_{t,j} := \bs{s}_{t-1}^{\top} \bs{W} \bs{h}_j$ as 
a bilinear transformation of encoder and decoder states, 
where $\bs{W}$ 
is a model parameter.%
\footnote{This is the default implementation in the OpenNMT  package. In preliminary experiments, 
feedforward attention \citep{bahdanau2014neural} did not show improvements.} %

\section{Sparse and Constrained Attention}\label{sec:attentions}

In this work, we consider alternatives to 
Eq.~\ref{eq:softmax_attention}. 
Since the softmax is strictly positive, it forces all words in the source to receive \emph{some} probability mass in the resulting attention distribution, which can be wasteful. 
Moreover, it may happen that the decoder attends repeatedly to the same source words across time steps, causing repetitions in the generated translation, as \citet{Tu:2016:ACL} observed.

With this in mind, we replace Eq.~\ref{eq:softmax_attention} by 
$\bs{\alpha}_t = \rho(\bs{z}_{t}, \bs{u}_{t})$,
where $\rho$ is a transformation that may depend both 
on the scores $\bs{z}_{t} \in \mathbb{R}^J$ and on 
{\bf upper bounds} $\bs{u}_t \in \mathbb{R}^J$ that limit the amount of attention that each word can receive. 
We consider three alternatives to softmax, described next. 

\begin{figure*}[t]
\includegraphics[width=\textwidth]{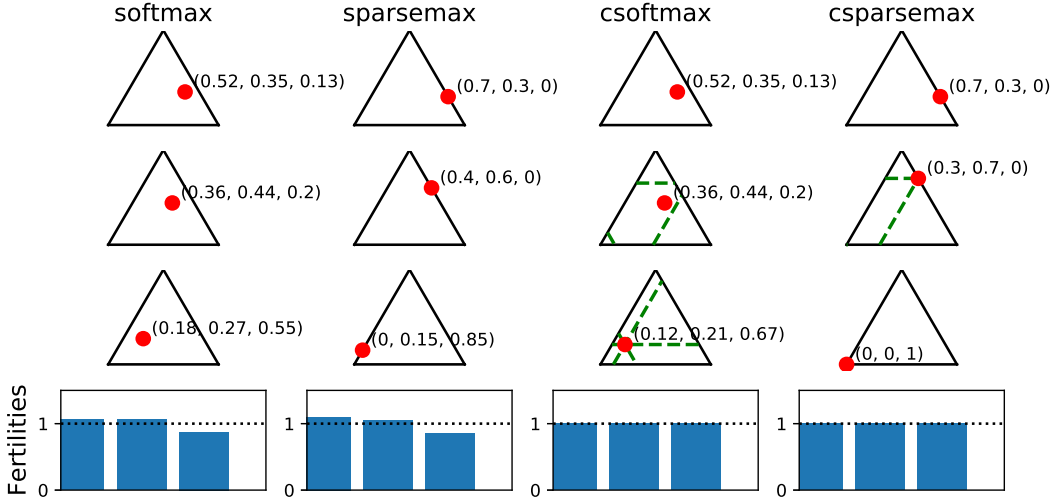}
\caption{Illustration of the different attention transformations for a toy example with three source words. We show the attention values on the probability simplex. In the first row we assume scores $\bs{z} = (1.2, 0.8, -0.2)$, and in the second and third rows $\bs{z} = (0.7, 0.9, 0.1)$ and $\bs{z} = (-0.2, 0.2, 0.9)$, respectively. For constrained softmax/sparsemax, we set unit fertilities to every word; for each row the upper bounds (represented as green dashed lines) are set as the difference between these fertilities and the cumulative attention each word has received. 
The last row illustrates the cumulative attention for the three words after all rounds.}
\end{figure*}

\paragraph{\bf Sparsemax.}

The sparsemax transformation 
\citep{martins2016softmax} 
is defined as: 
\begin{eqnarray}\label{eq:sparsemax}
\sparsemax(\bs{z}) 
&:=& \argmin_{\bs{\alpha} \in \Delta^{J}} \|\bs{\alpha} - \bs{z}\|^2,
\end{eqnarray}
where $\Delta^{J}:=\{\bs{\alpha} \in \mathbb{R}^J\,\,|\,\,\ \bs{\alpha} \ge \mathbf{0}, \sum_j \alpha_j = 1\}$.  
In words, it is the Euclidean projection of 
the scores $\bs{z}$ onto the probability simplex. 
These projections tend to hit the boundary of the simplex, yielding a sparse probability distribution. 
This 
allows the decoder to attend only to a few words in the source, assigning zero probability mass to all other words. 
\citet{martins2016softmax} have shown that the sparsemax can be evaluated in $O(J)$ time (same asymptotic cost as softmax) and 
gradient backpropagation takes sublinear time (faster than softmax), by exploiting the sparsity of the solution. 

\paragraph{\bf Constrained softmax.}

The constrained softmax transformation was recently proposed by 
\citet{martins2017learning} in the context of easy-first sequence tagging, being defined as follows:
\begin{eqnarray}\label{eq:csoftmax}
\csoftmax(\bs{z}; \bs{u}) &\!\!:=\!\!& 
\argmin_{\bs{\alpha} \in \Delta^{J}}  \mathsf{KL}(\bs{\alpha} \| \softmax(\bs{z}))\nonumber\\
&& \qquad \text{s.t.} \quad \bs{\alpha} \le \bs{u},
\end{eqnarray}
where $\bs{u}$ is a vector of upper bounds, 
and $\mathsf{KL}(.\|.)$ 
is the Kullback-Leibler divergence. 
In other words, it returns the 
distribution closest to $\softmax(\bs{z})$ 
whose attention probabilities are bounded by 
$\bs{u}$. 
\citet{martins2017learning} have shown that this transformation can be evaluated in $O(J\log J)$ time and 
its gradients backpropagated in $O(J)$ time. 

To use this transformation in the attention mechanism, 
we make use of the idea of {\bf fertility}  
\citep{Brown1993}. 
Namely, let $\bs{\beta}_{t-1} := \sum_{\tau=1}^{t-1} \bs{\alpha}_{\tau}$ denote the {\bf cumulative attention} that each source word has received up to time step $t$, 
and let $\bs{f} := (f_j)_{j=1}^J$ be a vector containing fertility upper bounds for each source word.  
The attention at step $t$ is  computed as 
\begin{equation}\label{eq:csoftmax_attention}
\bs{\alpha}_t = \csoftmax(\bs{z}_t, \bs{f}-\bs{\beta}_{t-1}).
\end{equation}
Intuitively, each source word $j$ gets a credit of 
$f_j$ units of attention, which are consumed along the decoding process. If all the credit is exhausted, 
it receives zero attention from then on. Unlike the sparsemax transformation, which places sparse attention over the source words, the constrained softmax leads to sparsity over time steps. 

\paragraph{\bf Constrained sparsemax.}

In this work, we propose a novel transformation 
which 
shares the two properties above: it
provides \emph{both} sparse and bounded probabilities. It is defined as:
\begin{eqnarray}\label{eq:csparsemax}
\csparsemax(\bs{z}; \bs{u}) 
&\!\!:=\!\!& \argmin_{\bs{\alpha} \in \Delta^{J}} \|\bs{\alpha} - \bs{z}\|^2 \nonumber\\
&& \qquad \text{s.t.} \quad \bs{\alpha} \le \bs{u}.
\end{eqnarray}
The following result, whose detailed proof we include as supplementary material (Appendix~\ref{sec:proof}), is key for enabling the use of the constrained sparsemax transformation in neural networks. 
\begin{proposition}\label{prop:csparsemax}
Let $\bs{\alpha}^\star = \csparsemax(\bs{z}; \bs{u})$ be the solution of 
Eq.~\ref{eq:csparsemax}, 
and define the sets 
$\mathcal{A} = \{j \in [J]\,\,|\,\, 0 < \alpha_j^\star < u_j\}$,
$\mathcal{A}_{L} = \{j \in [J]\,\,|\,\, \alpha_j^\star = 0\}$, 
and
$\mathcal{A}_{R} = \{j \in [J]\,\,|\,\, \alpha_j^\star = u_j\}$.  
Then: 
\smallskip
\begin{itemizesquish}
\item {\bf Forward propagation.} 
$\bs{\alpha}^\star$ can be computed
in $O(J)$ time with the algorithm of \citet{Pardalos1990} (Alg.~\ref{alg:pardalos} in Appendix~\ref{sec:proof}).
The solution takes the form $\alpha_j^\star =  \max\{0, \min \{u_j, z_j-\tau\}\}$, 
where $\tau$ 
is a normalization constant. 
\smallskip
\item {\bf Gradient backpropagation.} 
Backpropagation takes \emph{sublinear} time 
$O(|\mathcal{A}|+|\mathcal{A_R}|)$. 
Let $L(\bs{\theta})$ be a loss function, $\mathrm{d}\bs{\alpha} = \nabla_{\bs{\alpha}} L(\bs{\theta})$ be the output gradient, and 
$\mathrm{d}\bs{z} = \nabla_{\bs{z}} L(\bs{\theta})$ and $\mathrm{d}\bs{u} = \nabla_{\bs{u}} L(\bs{\theta})$ be the input gradients. 
Then, 
we have: 
\begin{eqnarray}
\mathrm{d}z_j &=& \mathds{1}(j \in \mathcal{A})(\mathrm{d}\alpha_j - m)\\
\mathrm{d}u_j &=& \mathds{1}(j \in \mathcal{A}_R)(\mathrm{d}\alpha_j - m),
\end{eqnarray}
where $m = \frac{1}{|\mathcal{A}|}\sum_{j \in \mathcal{A}} \mathrm{d}\alpha_j$. 
\end{itemizesquish}
\end{proposition}
%

\section{Fertility Bounds}\label{sec:fertilities}

We experiment with three ways of setting the fertility of the source words: \textsc{constant}, \textsc{guided}, and \textsc{predicted}. With \textsc{constant}, we set the fertilities of all source words to a fixed integer value $f$. With \textsc{guided}, we train a word aligner based on IBM Model 2 (we used {\tt fast\_align} in our experiments, \citet{Dyer2013}) and, for each word in the vocabulary, we set the fertilities to the maximal observed value in the training data (or 1 if no alignment was observed). With the \textsc{predicted} strategy, we train a separate fertility predictor model using a bi-LSTM tagger.%
\footnote{A similar strategy was recently used by \citet{Gu2018} as a component of their non-autoregressive NMT model.} %
At training time, we 
provide as supervision the fertility estimated by 
{\tt fast\_align}. 
Since our model works with fertility \emph{upper bounds} and the word aligner may miss some word pairs, 
we found it beneficial to add a constant to this number (1 in our experiments). 
At test time, we use the expected fertilities according to our model. 

\begin{figure}[t]
\centering
\includegraphics[scale=0.27,angle=90]{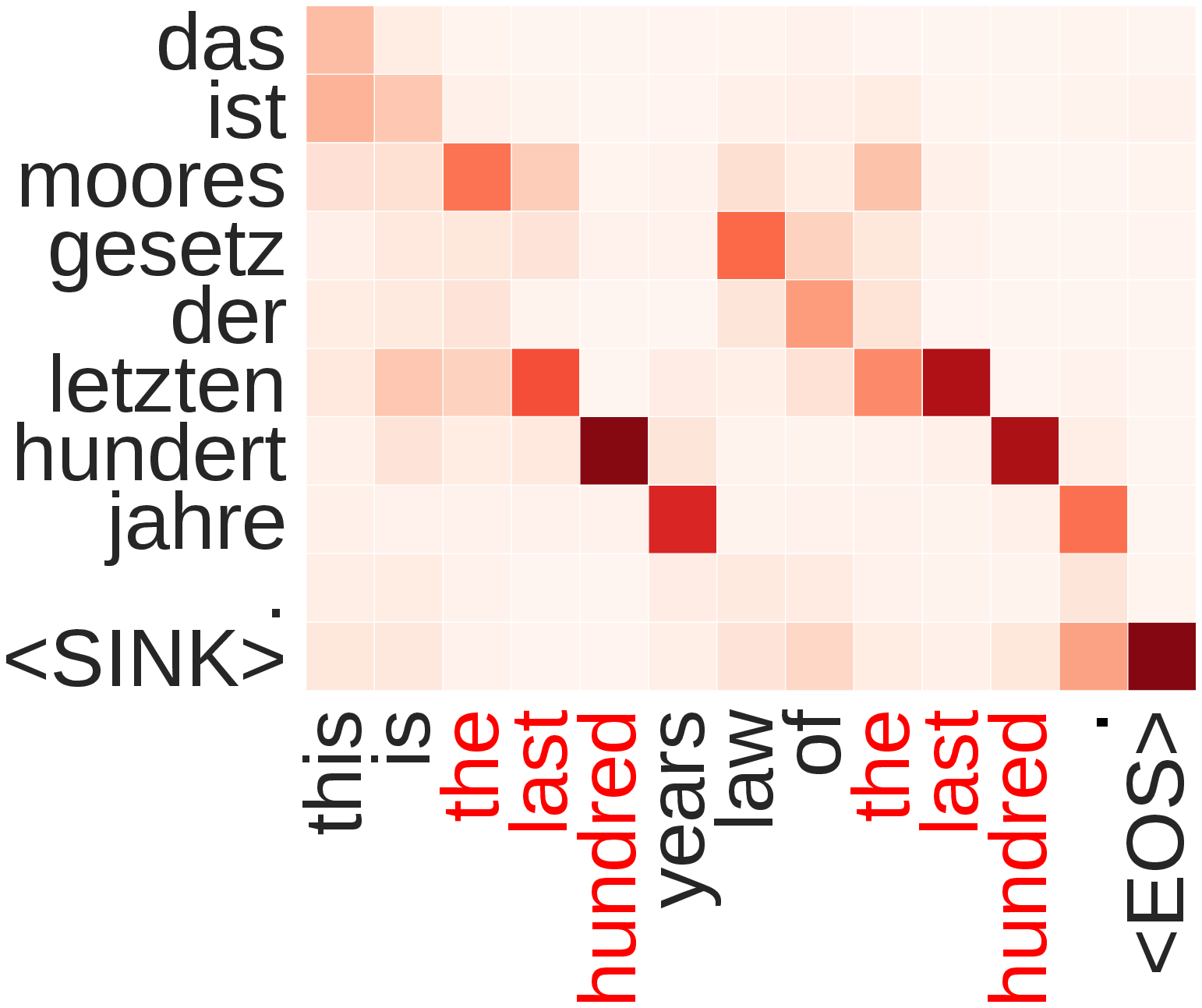}
\includegraphics[scale=0.27]{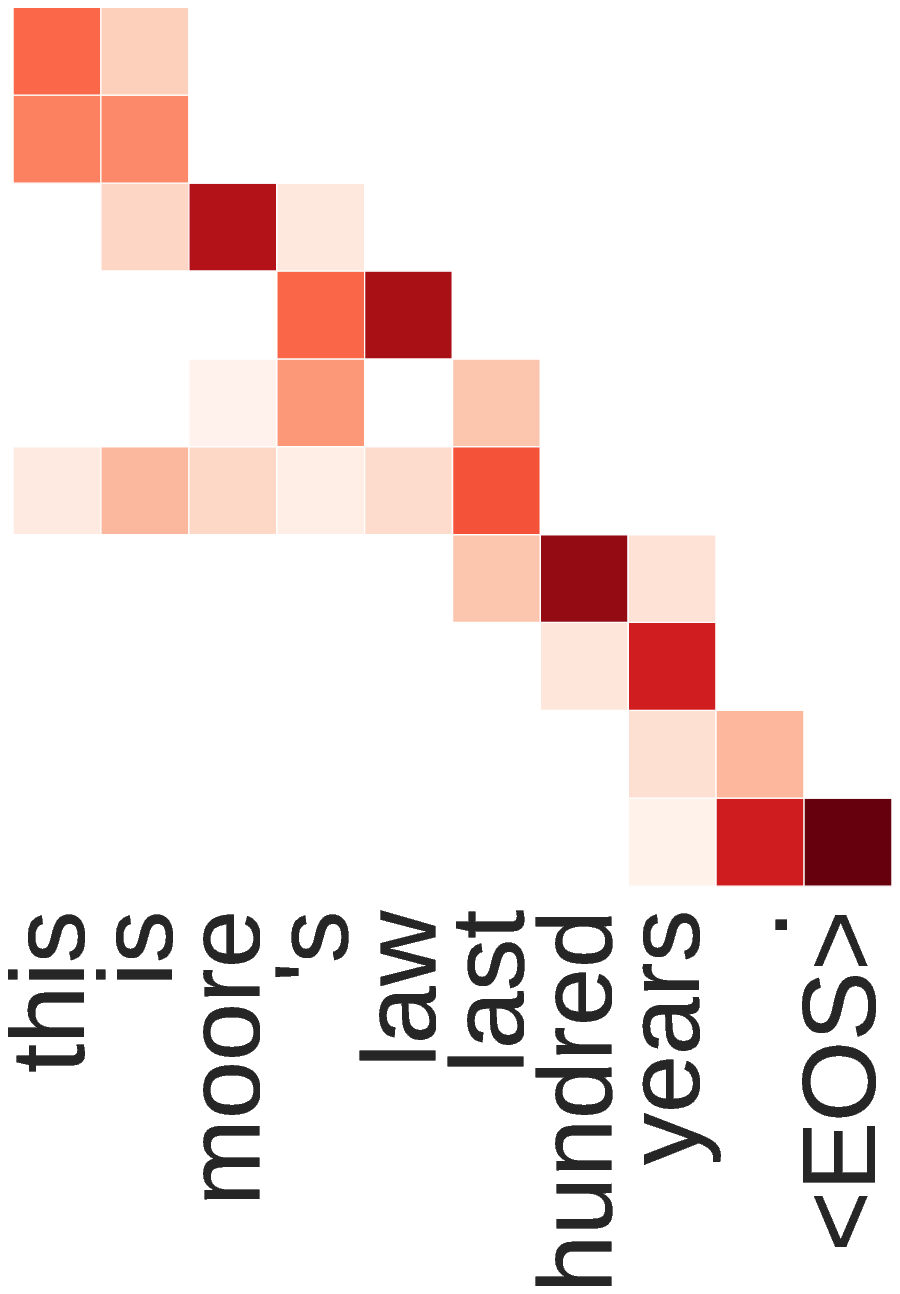}
\quad
\includegraphics[scale=0.35]{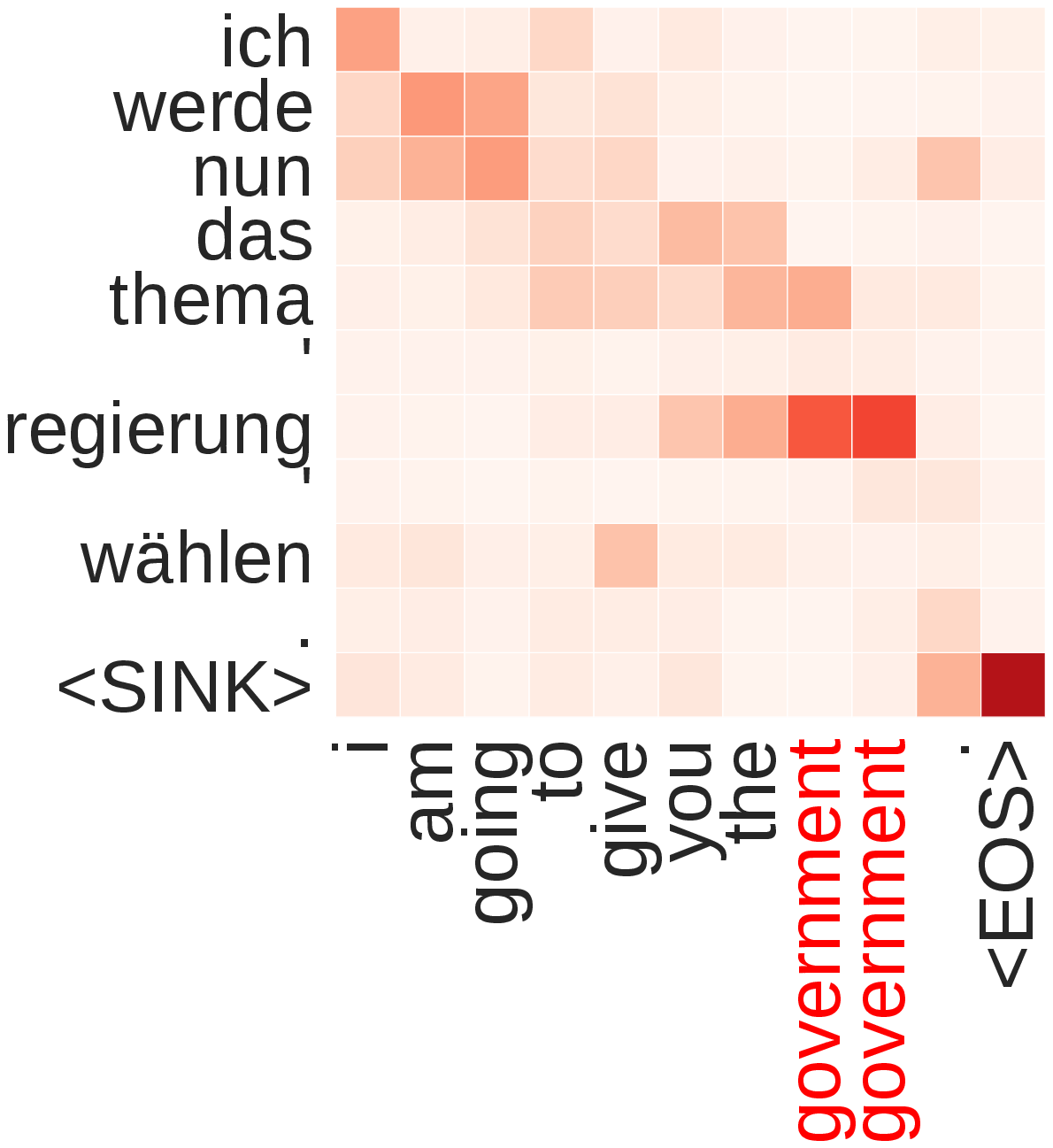}
\includegraphics[scale=0.35]{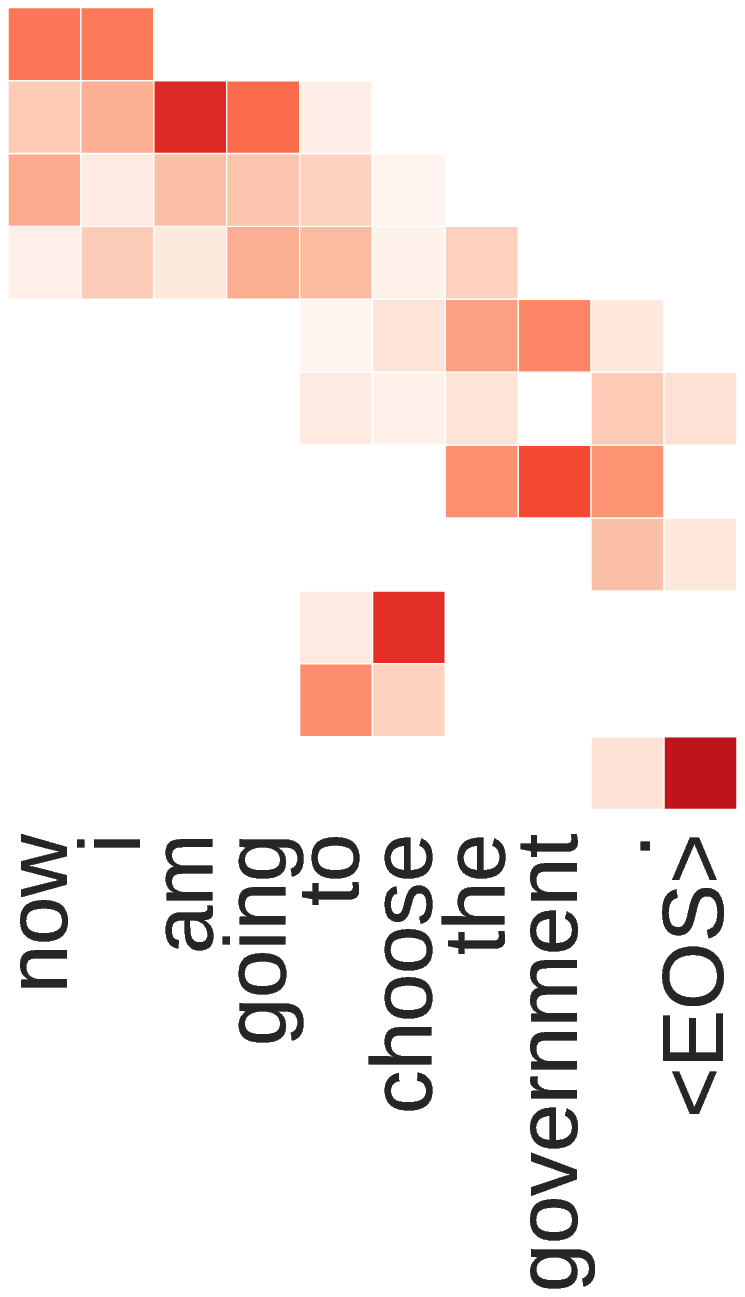}
\vspace{-0.3cm}
    \caption{Attention maps for $\softmax$ and $\csparsemax$ for two {\sc De-En} sentence pairs (white means zero attention). Repeated words are {\color{red}{highlighted}}. The reference translations are \emph{``This is Moore's law over the last hundred years''}  and \emph{``I am going to go ahead and select government.''}}%
    \label{fig:att_maps}%
\end{figure}

\begin{table*}[t!]
\begin{center}
\footnotesize
\begin{tabular}{@{\hskip 0in}l|c@{\hskip 0.05in}c@{\hskip 0.05in}c@{\hskip 0.1in}c|c@{\hskip 0.05in}c@{\hskip 0.05in}c@{\hskip 0.1in}c|c@{\hskip 0.05in}c@{\hskip 0.05in}c@{\hskip 0.1in}c@{\hskip 0in}}
& \multicolumn{4}{c|}{De-En} & \multicolumn{4}{c|}{Ja-En} & \multicolumn{4}{c}{Ro-En} \\ 
& {\sc bleu} & {\sc meteor} & {\sc rep} & {\sc drop} & {\sc bleu} & {\sc meteor} & {\sc rep} & {\sc drop} & {\sc bleu} & {\sc meteor} & {\sc rep} & {\sc drop} \\
\hline 
 $\softmax$ & 
 29.51 & 31.43 & 3.37 & 5.89 &
 20.36 & 23.83 & 13.48 & 23.30 &
 29.67 & 32.05 & 2.45 & 5.59 \\
 $\softmax$ + {\sc CovPenalty} & 
 29.69 & 31.53 & 3.47 & 5.74 &
 20.70 & 24.12 & 14.12 & 22.79 &
 29.81 & 32.15 & 2.48 & 5.49 \\
 $\softmax$ + {\sc CovVector} & 
 29.63 & 31.54 & 2.93 & 5.65 &
 21.53 & 24.50 & 11.07 & 22.18 &
 {\bf 30.08} & {\bf 32.22} & 2.42 & 5.47 \\
 $\sparsemax$ & 
 29.73 & 31.54 & 3.18 & 5.90 &
 21.28 & 24.25 & 13.09 & 22.40 &
 29.97 & 32.12 & 2.19 & 5.60 \\
 $\sparsemax$ + {\sc CovPenalty} & 
 29.83 & 31.60 & 3.24 & 5.79 &
 {\bf 21.64} & 24.49 & 13.36 & 21.91 &
 30.07 & 32.20 & 2.20 & 5.47 \\
 $\sparsemax$ + {\sc CovVector} & 
 29.22 & 31.18 & 3.13 & 6.15 &
 21.35 & {\bf 24.74} & {\bf 10.11} & {\bf 21.25} &
 29.30 & 31.84 & 2.18 & 5.87 \\
 $\csoftmax$ ($c=0.2$) & 
 29.39 & 31.33 & 3.29 & 5.86 &
 20.71 & 24.00 & 12.38 & 22.73 &
 29.39 & 31.83 & 2.37 & 5.64 \\
 $\csparsemax$ ($c=0.2$) & 
 {\bf 29.85} & {\bf 31.76} & {\bf 2.67} & {\bf 5.23} &
 21.31 & 24.51 & 11.40 & 21.59 &
 29.77 & 32.10 & {\bf 1.98} & {\bf 5.44} \\
\hline
\end{tabular}
\end{center}
\vspace{-0.3cm}
\caption{\label{tab:results} BLEU, METEOR, REP and DROP scores on the test sets for different attention transformations.}
\end{table*}

\begin{table}[t!]
\begin{center}
\small
\begin{tabular}{lc@{\hskip 0.1in}c}
& {\sc bleu} & {\sc meteor} \\
\hline
{\sc constant}, $f=2$ & 29.66 & 31.60 \\
{\sc constant}, $f=3$ & 29.64 & 31.56 \\
{\sc guided}, & 29.56 & 31.45 \\
{\sc predicted}, $c=0$ & 29.78 & 31.60 \\
{\sc predicted}, $c=0.2$ & {\bf 29.85} & {\bf 31.76} \\
\hline
\end{tabular}
\end{center}
\vspace{-0.3cm}
\caption{\label{tab:ablation} Impact of various fertility strategies for the $\csparsemax$ attention model ({\sc De-En}). }
\end{table}


\paragraph{Sink token.} We append an additional 
{$<${\sc sink}$>$}  
token to the end of the source sentence, to which we assign unbounded fertility ($f_{J+1} = \infty$). 
The token is akin to the null alignment in IBM models. 
The reason we add this token is the following: without the sink token, the length of the generated target sentence can never exceed $\sum_j f_j$ words if we use constrained softmax/sparsemax. %
At training time this may be problematic, since 
the target length is fixed and the problems in Eqs.~\ref{eq:csoftmax}--\ref{eq:csparsemax} can become infeasible. 
By adding the sink token we guarantee $\sum_j f_j = \infty$, eliminating the problem.

\paragraph{Exhaustion strategies.} To avoid missing source words, 
we implemented a simple strategy to encourage more attention to words with larger credit: 
we redefine the pre-attention word scores as
$\bs{z}_t' = \bs{z}_t + c \bs{u}_t$, 
where $c$ is a constant ($c=0.2$ in our experiments).
This increases the score of words which have not yet exhausted their fertility (we may regard it as a ``soft'' lower bound in Eqs.~\ref{eq:csoftmax}--\ref{eq:csparsemax}).  

\section{Experiments}

We evaluated our attention transformations on three language pairs. 
We focused on small datasets, as they are the most affected by 
coverage mistakes.  
We use the IWSLT 2014 corpus for \textsc{De-En}, 
the KFTT 
corpus for \textsc{Ja-En} \citep{neubig11kftt}, 
and the WMT 2016 dataset for \textsc{Ro-En}. 
The training sets have 153,326, 329,882, and 560,767 
parallel sentences, respectively. Our reason to prefer smaller datasets is that this regime is what brings more adequacy issues and demands more structural biases, hence it is a good test bed for our methods.
We tokenized 
the data using the Moses scripts and preprocessed it with subword units \cite{sennrich2016subword} with a joint vocabulary and 32k merge operations. 
	Our implementation was done on a fork of the OpenNMT-py toolkit \cite{klein2017} with the default parameters
\footnote{We used a 2-layer LSTM, embedding and hidden size of 500, dropout 0.3, and the SGD optimizer for 13 epochs.}. 
We used a validation set to tune hyperparameters introduced by our model. %
Even though our attention implementations are CPU-based using NumPy (unlike the rest of the computation which is done on the GPU), we did not observe any noticeable slowdown using multiple devices.

As baselines, we use softmax attention, as well as two recently proposed coverage models:
\begin{itemizesquish}
\item {\sc CovPenalty} \cite[\S{7}]{wu2016google}.    
At test time, the hypotheses in the beam are rescored with a global score that includes a length and a coverage penalty.%
\footnote{Since our sparse attention can become $0$ for some words, we extended the original coverage penalty by adding another parameter $\epsilon$, set to $0.1$: 
$\mathsf{cp}(x;y) := \beta \sum_{j=1}^J \log \max\{\epsilon, \min\{1, \sum_{t=1}^{|y|} \alpha_{jt}\}\}$.} 
We tuned $\alpha$ and $\beta$ with grid search on $\{0.2k\}_{k=0}^5$, as in \citet{wu2016google}.
\item {\sc CovVector} \citep{Tu:2016:ACL}. At training and test time, coverage vectors $\bs{\beta}$ and additional parameters $\bs{v}$ are used to condition the next attention step.  
We adapted this to our bilinear attention by defining 
$z_{t,j}= \bs{s}_{t-1}^{\top} (\bs{W} \bs{h}_j + \bs{v} {\beta}_{t-1,j})$. 
\end{itemizesquish}
We also experimented combining the strategies above with the sparsemax transformation. 

As evaluation metrics, we report tokenized BLEU, 
METEOR  (\citet{denkowski:lavie:meteor-wmt:2014}, 
as well as two new metrics that we describe next to account for over and under-translation.%
\footnote{Both evaluation metrics are included in our software package at \texttt{www.github.com/Unbabel/\\sparse\_constrained\_attention}.} %

\paragraph{REP-score:} a new metric to count repetitions.
Formally, given an $n$-gram $s \in V^n$, let $t(s)$ and $r(s)$ be the its frequency in the model translation and reference. 
We first compute a sentence-level score
\begin{eqnarray*}\label{eq:rep_score}
\sigma(t,r) &=& \textstyle \lambda_1 \sum_{s \in V^n, \,\, t(s) \ge 2} \max\{0, t(s) - r(s)\} \nonumber\\
&+&\textstyle \lambda_2 \sum_{w \in V} \max\{0, t(ww) - r(ww)\}.
\end{eqnarray*}
The REP-score is then given by summing $\sigma(t,r)$ over sentences, normalizing by the number of words on the reference corpus, and multiplying by 100. We used $n=2$, $\lambda_1 = 1$ and $\lambda_2 = 2$. 

\paragraph{DROP-score:} a new metric that accounts for possibly dropped words. 
To compute it, we first compute two sets of word alignments: from source to reference translation, and 
from source to the predicted translation. In our experiments, the alignments were obtained with \texttt{fast\_align} \cite{Dyer2013}, trained on the training partition of the data. 
Then, the DROP-score computes the percentage of source words that aligned with some word from the reference translation, but not with any word from the predicted translation. 

\bigskip




Table~\ref{tab:results} shows the results. We can see that on average, the sparse models ($\csparsemax$ as well as $\sparsemax$ combined with coverage models) have higher scores on both BLEU and METEOR. Generally, they also obtain better REP and DROP scores than $\csoftmax$ and $\softmax$, which suggests that sparse attention alleviates the problem of coverage to some extent. 

To compare different fertility strategies, we ran experiments on the {\sc De-En} for the $\csparsemax$ transformation (Table~\ref{tab:ablation}). We see that the {\sc Predicted} strategy outperforms the others both in terms of BLEU and METEOR, albeit slightly.

Figure \ref{fig:att_maps} 
shows examples of sentences for which the $\csparsemax$ fixed repetitions, along with the corresponding attention maps. 
We see that in the case of $\softmax$ repetitions, the decoder attends repeatedly to the same portion of the source sentence (the expression \textit{``letzten hundert"} in the first sentence and  \textit{``regierung"} in the second sentence).  
Not only did $\csparsemax$ avoid repetitions, 
but it also yielded a sparse set of alignments, as expected. Appendix~\ref{sec:translations} provides more examples of translations from all models in discussion.

\section{Conclusions}

We proposed a new approach to address the coverage problem in NMT,  
by replacing the softmax attentional transformation by sparse and constrained alternatives: sparsemax, 
constrained softmax, and the newly proposed constrained sparsemax. 
For the latter, we derived efficient forward and backward propagation algorithms. 
By incorporating a model for fertility prediction, our attention transformations led to sparse alignments, avoiding repeated words in the translation.

\section*{Acknowledgments}

We thank the Unbabel AI Research team for numerous discussions, and the three anonymous reviewers for their insightful 
comments.
This work was 
supported by the European Research Council (ERC
StG DeepSPIN 758969) 
and by the Funda\c{c}\~ao para a Ci\^encia e Tecnologia through contracts UID/EEA/50008/2013, PTDC/EEI-SII/7092/2014 (LearnBig), and CMUPERI/TIC/0046/2014 ~(GoLocal).

\bibliography{acl2018}
\bibliographystyle{acl_natbib}

\appendix
\newpage
\onecolumn

\section{Proof of Proposition~\ref{prop:csparsemax}}
\label{sec:proof}

We provide here a detailed proof of Proposition~\ref{prop:csparsemax}. 

\subsection{Forward Propagation}

The optimization problem can be written as
\begin{eqnarray}
\csparsemax(\bs{z},\bs{u}) = \argmin & \frac{1}{2}\|\bs{\alpha}\|^2 - \DP{\bs{z}}{\bs{\alpha}} \nonumber\\
\text{s.t.} & 
\left\{
\begin{array}{l}
\DP{\mathbf{1}}{\bs{\alpha}} = 1\nonumber\\
\mathbf{0} \le \bs{\alpha} \le \bs{u}.
\end{array}
\right.
\end{eqnarray}
The Lagrangian function is:
\begin{eqnarray}
\mathcal{L}(\bs{\alpha}, \tau, \bs{\mu}, \bs{\nu}) = 
-\frac{1}{2}\|\bs{\alpha}\|^2 - \DP{\bs{z}}{\bs{\alpha}} + \tau(\DP{\mathbf{1}}{\bs{\alpha}} - 1) 
 -\DP{\bs{\mu}}{\bs{\alpha}} + \DP{\bs{\nu}}{(\bs{\alpha} - \bs{u})}.
\end{eqnarray}
To obtain the solution, we invoke the Karush-Kuhn-Tucker conditions. 
From the stationarity condition, we have 
$\mathbf{0} = \bs{\alpha} - \bs{z} + \tau \mathbf{1} - \bs{\mu} + \bs{\nu}$, 
which due to the primal feasibility condition implies that the solution is of the form:
\begin{equation}
\bs{\alpha} = \bs{z} - \tau \mathbf{1} + \bs{\mu} - \bs{\nu}.
\end{equation}
From the complementarity slackness condition, 
we have that $0 < \alpha_j < u_j$ implies that $\mu_j = \nu_j = 0$ and therefore $\alpha_j = z_j - \tau$. 
On the other hand, $\mu_j > 0$ implies $\alpha_j = 0$, and $\nu_j > 0$ implies $\alpha_j = u_j$. 
Hence the solution can be written as $\alpha_j = \max\{0, \min \{u_j, z_j - \tau\}$, 
where $\tau$ is determined such that the distribution normalizes:
\begin{equation}\label{eq:csparsemax_constant}
\tau = \frac{\sum_{j\in \mathcal{A}} z_j + \sum_{j \in \mathcal{A}_{R}} u_j - 1}{|\mathcal{A}|},
\end{equation}
with $\mathcal{A} = \{j \in [J]\,\,|\,\, 0 < \alpha_j < u_j\}$ and 
$\mathcal{A}_{R} = \{j \in [J]\,\,|\,\, \alpha_j = u_j\}$.  
Note that $\tau$ depends itself on the set $\mathcal{A}$, a function of the solution. 
In \S\ref{sec:pardalos}, we describe an algorithm that searches the value of $\tau$ efficiently.

\subsection{Gradient Backpropagation}

We now turn to the problem of backpropagating the gradients through the 
constrained sparsemax transformation. 
For that, we need to compute its Jacobian matrix, i.e., 
the derivatives $\frac{\partial \alpha_i}{\partial z_j}$ and $\frac{\partial \alpha_i}{\partial u_j}$ for $i,j\in [J]$. 
Let us first express $\bs{\alpha}$ as  
\begin{equation}\label{eq:csoftmax_branches}
\alpha_i = \left\{
\begin{array}{ll}
0, & i \in \mathcal{A}_L,\\
z_i - \tau, & i \in \mathcal{A},\\
u_i, & i \in \mathcal{A}_R,
\end{array}
\right.
\end{equation}
with $\tau$ as in Eq.~\ref{eq:csparsemax_constant}. 
Note that we have ${\partial \tau}/{\partial z_j} = \mathds{1}(j \in \mathcal{A}) / |\mathcal{A}|$ and 
${\partial \tau}/{\partial u_j} = \mathds{1}(j \in \mathcal{A}_R) / |\mathcal{A}|$. 
Thus, we have the following:
\begin{equation}
\frac{\partial \alpha_i}{\partial z_j} = \left\{
\begin{array}{ll}
1-1/|\mathcal{A}|, & \text{if $j \in \mathcal{A}$ and $i=j$}\\
-1/|\mathcal{A}|, & \text{if $i,j \in \mathcal{A}$ and $i\ne j$}\\
0, & \text{otherwise,}
\end{array}
\right.
\end{equation}
and
\begin{equation}
\frac{\partial \alpha_i}{\partial u_j} = \left\{
\begin{array}{ll}
1, & \text{if $j \in \mathcal{A}_R$ and $i=j$}\\
-1/|\mathcal{A}|, & \text{if $j \in \mathcal{A}_R$ and $i\in \mathcal{A}$}\\
0, & \text{otherwise.}
\end{array}
\right.
\end{equation}
Finally, we obtain:
\begin{eqnarray}
\mathrm{d}z_j &=& \sum_{i} \frac{\partial \alpha_i}{\partial z_j} \mathrm{d}\alpha_i \nonumber\\
&=& \mathds{1}(j\in \mathcal{A}) \left(\mathrm{d}\alpha_j - \frac{\sum_{i \in \mathcal{A}} \mathrm{d}\alpha_i}{|\mathcal{A}|}\right) \nonumber\\
&=& \mathds{1}(j\in \mathcal{A}) (\mathrm{d}\alpha_j - m),
\end{eqnarray} 
and
\begin{eqnarray}
\mathrm{d}u_j &=& \sum_{i} \frac{\partial \alpha_i}{\partial u_j} \mathrm{d}\alpha_i \nonumber\\
&=& \mathds{1}(j\in \mathcal{A}_R) \left(\mathrm{d}\alpha_j - \frac{\sum_{i \in \mathcal{A}} \mathrm{d}\alpha_i}{|\mathcal{A}|}\right) \nonumber\\
&=& \mathds{1}(j \in \mathcal{A}_R) (\mathrm{d}\alpha_j - m),
\end{eqnarray} 
where $m = \frac{1}{|\mathcal{A}|}\sum_{j \in \mathcal{A}} \mathrm{d}\alpha_j$.

\begin{algorithm}[t]
{ %
 \caption{{Pardalos and Kovoor's Algorithm}\label{alg:pardalos}}
\begin{algorithmic}[1]
  \STATE {\bfseries input:} 
   $\bs{a}, \bs{b}, \bs{c}, d$
   \STATE Initialize working set $\mathcal{W}\leftarrow \{1,\ldots,J\}$
   \STATE Initialize set of split points:\label{algline:splitpoints} 
   $$\mathcal{P}\leftarrow \{a_j, b_j\}_{j=1}^J \cup \{\pm \infty\}$$
   \STATE Initialize $\tau_{\mathrm{L}}\leftarrow -\infty$, 
    $\tau_{\mathrm{R}}\leftarrow \infty$, $s_{\mathrm{tight}}\leftarrow 0$, $\xi \leftarrow 0$.
	\WHILE{$\mathcal{W} \ne \varnothing$}
	\STATE Compute $\tau\leftarrow \mathrm{Median}(\mathcal{P})$
	\STATE Set $s \leftarrow s_{\mathrm{tight}} + 
\sum_{j\in \mathcal{W}\,|\, b_i < \tau} c_j b_j + \sum_{j\in \mathcal{W}\,|\, a_j > \tau} c_j a_j + (\xi + \sum_{j \in \mathcal{W}\,|\, a_j \le \tau \le b_j} c_j)\tau$ 	
	\STATE If $s \le d$, set $\tau_{\mathrm{L}} \leftarrow \tau$;
	if $s \ge d$, set $\tau_{\mathrm{R}} \leftarrow \tau$
	\STATE Reduce set of split points: $\mathcal{P} \leftarrow \mathcal{P} \cap [\tau_{\mathrm{L}},\tau_{\mathrm{R}}]$
	\STATE Update tight-sum: $s_{\mathrm{tight}} \leftarrow s_{\mathrm{tight}} + 
\sum_{j\in \mathcal{W}\,|\, b_i < \tau_L} c_j b_j + \sum_{j\in \mathcal{W}\,|\, a_j > \tau_R} c_j a_j$
	\STATE Update slack-sum: $\xi \leftarrow \xi + \sum_{j\in \mathcal{W}\,|\,a_j \le \tau_L \wedge b_j \ge \tau_R} c_j$
    \STATE Update working set: $\mathcal{W} \leftarrow \{j \in \mathcal{W} \,|\, \tau_L < a_j < \tau_R \vee \tau_L < b_j < \tau_R\}$
	\ENDWHILE
   \STATE Define $y^* \leftarrow(d-s_{\mathrm{tight}})/\xi$
   \STATE Set $x_j^{\star} = \max\{a_j, \min\{b_j, y\}\}, \,\, \forall j \in [J]$
   \STATE \textbf{output:} $\bs{x}^{\star}$. 
\end{algorithmic}}
\end{algorithm}

\subsection{Linear-Time Evaluation}\label{sec:pardalos}

Finally, we present an algorithm to solve the problem in Eq.~\ref{eq:csparsemax} in linear time. 

\citet{Pardalos1990} describe an algorithm, reproduced here as Algorithm~\ref{alg:pardalos}, for solving a class of singly-constrained convex quadratic problems, which can be written in the form above (where each $c_j\ge 0$):
\begin{align}\label{eq:reduced_problem}
\mathrm{minimize} \,\,& \sum_{j=1}^J c_j x_j^2\nonumber\\
\mathrm{s.t.} \,\, & \textstyle \sum_{j=1}^J c_j x_j = d,\nonumber\\
& \textstyle a_j \le x_j \le b_j, \quad j=1,\ldots,J. 
\end{align}
The solution of the problem in Eq.~\ref{eq:reduced_problem} is of the form
$x_j^{\star} = \max\{a_j, \min\{b_j, y\}\}$, where 
$y \in [a_j, b_j]$ is a constant.  
The algorithm searches the value of this constant (which is similar to $\tau$ in our problem), which lies in a particular interval of split-points (line~\ref{algline:splitpoints}), iteratively shrinking this interval.  
The algorithm requires computing medians as a subroutine, 
which can be done in linear time \citep{Blum1973}. 
The overall complexity in $O(J)$ \citep{Pardalos1990}. 
The same algorithm has been used in NLP by \citet{Almeida2013ACL} for a budgeted summarization problem. 

To show that this algorithm applies to the problem of evaluating $\csparsemax$, it suffices to show that our problem in Eq.~\ref{eq:csparsemax} can be rewritten in the form of Eq.~\ref{eq:reduced_problem}. 
This is indeed the case, if we set:
\begin{eqnarray}
x_j &=& \frac{p_j - z_j}{2}\\
a_j &=& -z_j / 2\\
b_j &=& (u_j-z_j)/2\\
c_j &=& 1\\
d &=& \frac{1 - \sum_{j=1}^J z_j}{2}.
\end{eqnarray}

\section{Examples of Translations}
\label{sec:translations}

We show some examples of translations obtained for the German-English language pair with different systems. \correct{Blue} highlights the parts of the reference that are correct and \wrong{red} highlights the corresponding problematic parts of translations, including repetitions, dropped words or mistranslations. 

\begin{table*}[h!]
\small
\begin{tabular}{|l|l|}
\hline
\textbf{input}             & \source{{\"u}berlassen sie das ruhig uns .} \\ \hline
\textbf{reference}         & \correct{leave that up to us} .          \\ \hline
$\softmax$    			   & \wrong{give us a silence} .            \\ \hline
$\sparsemax$			   & leave that to us .				\\ \hline
$\csoftmax$   			   & \wrong{let's} leave that .             \\ \hline
$\csparsemax$ 			   & leave it to us .               \\ \hline
\end{tabular}
\end{table*}


\begin{table*}[h!]
\small
\begin{tabular}{|l|l|}
\hline
\textbf{input}             & \source{so ungef{\"a}hr , sie wissen schon .} \\ \hline
\textbf{reference}         & \correct{like that , you know} .            \\ \hline
$\softmax$    			   & \wrong{so , you know , you know} . \\ \hline
$\sparsemax$			   & \wrong{so , you know , you know} .				\\ \hline
$\csoftmax$   			   & \wrong{so , you know , you know} .\\ \hline
$\csparsemax$ 			   & like that , you know .               \\ \hline
\end{tabular}
\end{table*}


\begin{table*}[h!]
\small
\begin{tabular}{|l|l|}
\hline
\textbf{input}     & \source{und wir benutzen dieses wort mit solcher verachtung .}  \\ \hline
\textbf{reference} & and we say that word \correct{with such contempt} .             \\ \hline
$\softmax$         & and we use this word with such \wrong{contempt contempt} .      \\ \hline
$\sparsemax$       & and we use this word with such contempt .			             \\ \hline
$\csoftmax$        & and we use this word with \wrong{like this} .                   \\ \hline
$\csparsemax$      & and we use this word with such contempt .                       \\ \hline
\end{tabular}
\end{table*}


\begin{table*}[h!]
\small
\begin{tabular}{|l|l|}
\hline
\textbf{input}     & \source{wir sehen das dazu , dass phosphor wirklich kritisch ist .}  \\ \hline
\textbf{reference} & we can see \correct{that} phosphorus is really critical .            \\ \hline
$\softmax$         & we see \wrong{that that} phosphorus is really critical .             \\ \hline
$\sparsemax$       & we see \wrong{that that} phosphorus really is critical .			  \\ \hline
$\csoftmax$        & we see \wrong{that that} phosphorus is really critical .             \\ \hline
$\csparsemax$      & we see that phosphorus is really critical .                          \\ \hline
\end{tabular}
\end{table*}


\begin{table*}[h!]
\small
\begin{tabular}{|l|l|}
\hline
\textbf{input}     &  \source{also m{\"u}ssen sie auch nicht auf klassische musik verzichten , weil sie kein instrument spielen .} \\ \hline
\textbf{reference} &  so \correct{you don't need to abstain from listening to} classical music because \correct{you don't play} an instrument . \\ \hline
$\softmax$         &  so you don't have to \wrong{rely on} classical music because you don't \wrong{have} an instrument .  \\ \hline
$\sparsemax$       & so \wrong{they} don't have to \wrong{kill} classical music because \wrong{they} don't play an instrument .		   \\ \hline
$\csoftmax$        & so \wrong{they} don't have to \wrong{rely on} classical music , because \wrong{they} don't play an instrument . \\ \hline
$\csparsemax$      & so you don't have to get \wrong{rid of} classical music , because you don't play an instrument .  \\ \hline
\end{tabular}
\end{table*}

\clearpage
\begin{table*}[h!]
\small
\begin{tabular}{|l|l|}
\hline
\textbf{input}     & \source{je mehr ich aber darüber nachdachte , desto mehr kam ich zu der ansicht , das der fisch etwas wei{\ss} .} \\ \hline
\textbf{reference} & the more i thought about it , however , the more \correct{i came to the view that this fish knows something }. \\ \hline
$\softmax$         & the more i thought about it , the more \wrong{i got to the point of the fish} .                              \\ \hline
$\sparsemax$	   & the more i thought about it , the more \wrong{i got to the point of view of the fish} .                      \\ \hline
$\csoftmax$        & but the more i thought about it , the more \wrong{i came to mind , the fish} .                               \\ \hline
$\csparsemax$      & the more i thought about it , the more i\wrong{ came to the point that the fish knows} .                     \\ \hline
\end{tabular}
\end{table*}


\begin{table*}[h!]
\small
\begin{tabular}{|l|p{0.85\textwidth}|}
\hline
\textbf{input}     & \source{all diese menschen lehren uns , dass es noch andere existenzm{\"o}glichkeiten , andere denkweisen , andere wege zur orientierung auf der erde gibt .}                                                                                 \\ \hline
\textbf{reference} & all of these peoples teach us that there are \correct{other ways of being , other ways of thinking , other ways of orienting yourself in the earth .}                                                                                             \\ \hline
$\softmax$         & all of these people teach us that there are \wrong{others , other ways , other ways of guidance to the earth .}               \\ \hline
$\sparsemax$       & all these people are teaching us that there are other \wrong{options , other ways , different} ways of guidance on earth .    \\ \hline
$\csoftmax$        & all of these people teach us that there's other ways of \wrong{doing} other ways of thinking , other ways of guidance on earth . \\ \hline
$\csparsemax$      & all these people teach us that there are other \wrong{actors} , other ways of thinking , other ways of guidance on earth .    \\ \hline
\end{tabular}
\end{table*}


\begin{table*}[h!]
\small
\begin{tabular}{|l|l|}
\hline
\textbf{input}     & \source{in der reichen welt , in der oberen milliarde , k{\"o}nnten wir wohl abstriche machen und weniger nutzen ,}              				  \\ 
                   & \source{aber im durchschnitt wird diese zahl jedes jahr steigen und sich somit insgesamt mehr als verdoppeln ,}              				  \\
                   & \source{die zahl der dienste die pro person bereitgestellt werden .}                                                         				  \\ \hline
\textbf{reference} & in the rich world , perhaps the top one billion , we probably \correct{could cut back and use less} , but every year , this number ,         \\
                   & on average , is going to go up , \correct{and so} , over all , that will \correct{more than double the services delivered per person} .      \\ \hline
$\softmax$         & in the rich world , in the upper billion , we might \wrong{be able to do} and \wrong{use} less \wrong{use} , but on average , that number    \\
                   & is going to increase every year and so \wrong{on} , \wrong{which is} the number of services \wrong{that are being put in .}          			      \\ \hline
$\sparsemax$       & in the rich world , in the upper billion , we may \wrong{be able to do} and \wrong{use} less \wrong{use} , but in average , that number	  \\
                   & is going to rise every year , and so \wrong{much} more than double , the number of services that are being \wrong{put together} . 			  \\ \hline
$\csoftmax$        & in the rich world , in the upper billion , we might \wrong{be able to take off} and use less , but in average , this number                  \\
                   & is going to increase every year and so \wrong{on} , and that's the \wrong{number of people who} are being put together per person .          \\ \hline
$\csparsemax$      & in the rich world , in the upper billion , we may \wrong{be able to turn off} and use less , but in average , that number will 			  \\
                   & rise every year and so far more than double , the number of services that are being put into a person . 									  \\ \hline
\end{tabular}
\end{table*}

\end{document}